\documentclass{ieeeaccess}
\usepackage{cite}
\usepackage{amsmath,amssymb,amsfonts}
\usepackage{algorithmic}
\usepackage{graphicx}
\usepackage{textcomp}
\usepackage{amsmath}
\usepackage{subcaption}
\usepackage{booktabs}
\usepackage{multirow}
\usepackage{url}

\newcommand{\vt}[1]{{\boldsymbol #1}}
\def\BibTeX{{\rm B\kern-.05em{\sc i\kern-.025em b}\kern-.08em
    T\kern-.1667em\lower.7ex\hbox{E}\kern-.125emX}}
\begin{document}
\history{Date of publication xxxx 00, 0000, date of current version xxxx 00, 0000.}
\doi{10.1109/ACCESS.2021.DOI}

\title{Disease-oriented image embedding with pseudo-scanner standardization for content-based image retrieval on 3D brain MRI}
%\author{\uppercase{First A. Author}\authorrefmark{1}, \IEEEmembership{Fellow, IEEE},
%\uppercase{Second B. Author\authorrefmark{2}, and Third C. Author,
%Jr}.\authorrefmark{3},
%\IEEEmembership{Member, IEEE}}
\author{\uppercase{Hayato Arai}\authorrefmark{1},
\uppercase{Yuto Onga\authorrefmark{1}, Kumpei Ikuta\authorrefmark{1}, Yusuke Chayama\authorrefmark{1}, Hitoshi Iyatomi\authorrefmark{1} \IEEEmembership{Member, IEEE}, and Kenichi Oishi} \authorrefmark{2} for the Alzheimer's Disease Neuroimaging Initiative and the Parkinson's Progression Markers Initiative.}
\address[1]{Department of Applied Informatics, Graduate School of Science and Engineering, Hosei University, Tokyo, 184-8584 Japan (e-mail: iyatomi@hosei.ac.jp)}
\address[2]{Department of Radiology and Radiological Science, Johns Hopkins University School of Medicine, Baltimore, MD 21205 USA (email:koishi2@jhmi.edu)} 
%CO 80523 USA (e-mail: author@lamar.colostate.edu)}
%\address[3]{Electrical Engineering Department, University of Colorado, Boulder, CO 
%80309 USA}
%\tfootnote{This paragraph of the first footnote will contain support 
%information, including sponsor and financial support acknowledgment. For 
%example, ``This work was supported in part by the U.S. Department of 
%Commerce under Grant BS123456.''}
\tfootnote{Data used in preparation of this article were obtained from the Alzheimer’s Disease Neuroimaging Initiative (ADNI) database (adni.loni.usc.edu). As such, the investigators within the ADNI contributed to the design and implementation of ADNI and/or provided data but did not participate in analysis or writing of this report. A complete listing of ADNI investigators can be found at: 
\url{http://adni.loni.usc.edu/wp-content/uploads/how_to_apply/ADNI_Acknowledgement_List.pdf}
}

\markboth
{Arai \headeretal: Preparation of Papers for IEEE TRANSACTIONS and JOURNALS}
{Arai \headeretal: Preparation of Papers for IEEE TRANSACTIONS and JOURNALS}

\corresp{Corresponding author: Hitoshi Iyatomi (e-mail: iyatomi@hosei.ac.jp).}

\begin{abstract}
To build a robust and practical content-based image retrieval (CBIR) system that is applicable to a clinical brain MRI database, we propose a new framework -- Disease-oriented image embedding with pseudo-scanner standardization (DI-PSS) -- that consists of two core techniques, data harmonization and a dimension reduction algorithm. Our DI-PSS uses skull stripping and CycleGAN-based image transformations that map to a standard brain followed by transformation into a brain image taken with a given reference scanner. Then, our 3D convolutioinal autoencoders (3D-CAE) with deep metric learning acquires a low-dimensional embedding that better reflects the characteristics of the disease. The effectiveness of our proposed framework was tested on the T1-weighted MRIs selected from the Alzheimer's Disease Neuroimaging Initiative and the Parkinson's Progression Markers Initiative. We confirmed that our PSS greatly reduced the variability of low-dimensional embeddings caused by different scanner and datasets. Compared with the baseline condition, our PSS reduced the variability in the distance from Alzheimer's disease (AD) to clinically normal (CN) and Parkinson disease (PD) cases by 15.8--22.6\% and 18.0--29.9\%, respectively. These properties allow DI-PSS to generate lower dimensional representations that are more amenable to disease classification. In AD and CN classification experiments based on spectral clustering, PSS improved the average accuracy and  macro-F1 by 6.2\% and 10.7\%, respectively. Given the potential of the DI-PSS for harmonizing images scanned by MRI scanners that were not used to scan the training data, we expect that the DI-PSS is suitable for application to a large number of legacy MRIs scanned in heterogeneous environments.
\end{abstract}


\titlepgskip=-15pt

\maketitle

\section{Introduction}
\label{sec:introduction}
\PARstart{I}{n} the new era of Open Science \cite{Woelfe2011}, data sharing has become increasingly crucial for efficient and fair development of science and industry. Especially in the field of medical image science, various datasets have been released and used for the development of new methods and benchmarks. There have been attempts to create publicly open databases consisting of medical images, demographic data, and clinical information, such as ADNI, AIBL, PPMI, 4RTN, PING, ABCD and UK BioBank. In the near future, clinical images acquired with medical indications will become available for research use. 

% needs for information retrieval, CBIR
Big data, consisting of large amounts of brain magnetic resonance (MR) images and corresponding medical records, could provide new evidence for the diagnosis and treatment of various diseases. 
Clearly, search technology is essential for the practical and effective use of such big data.
Currently, text-based searching is widely used for the retrieval of brain MR images. However, since this approach requires skills and experience during retrieval and data registration, there is a strong demand from the field to realize content-based image retrieval (CBIR) \cite{Kumar2013}.

To build a CBIR system that is feasible for brain MR imaging (MRI) databases, obtaining an appropriate and robust low-dimensional representation of the original MR images that reflects the characteristics of the disease in focus is extremely important.
Various methods have been proposed, including those based on classical feature description
%\cite{Tu2010}\cite{Huang2012}\cite{Murala2014},
\cite{Tu2010,Huang2012,Murala2014}, 
anatomical phenotypes \cite{Faria2015}, and deep learning techniques
\cite{Kruthika2019,Swati2019,Onga2019}. %\cite{Kruthika2019}\cite{Swati2019}\cite{Onga2019}.
The latter two techniques %\cite{Swati2019}\cite{Onga2019} 
\cite{Swati2019,Onga2019} acquire similar low-dimensional representations for similar disease data by introducing the idea of distance metric learning \cite{Alipanahi2008}\cite{Hoffer2016}.
Their low-dimensional representations adequately capture disease characteristics rather than individual variations seen on gyrification patterns in the brain. However, the application of these methods to a heterogeneous database containing MRIs from various scanners and scan protocols is hampered by the scanner or protocol bias, which is not negligible.

% Data harmonization（DH）intro
In brain MRI, such non-biological experimental variations (i.e., magnetic field strength, scanner manufacturer, reconstruction method) resulting from differences in scanner characteristics and protocols can affect the images in various ways and have a significant impact on the subsequent process
\cite{Clark2006,Han2006,Yu2018,Oishi2019,Onga2019,Wachinger2021}.
%\cite{Clark2006}\cite{Han2006}\cite{Yu2018}\cite{Oishi2019}\cite{Onga2019}\cite{Wachinger2021}.
%[Clark2005, Han2006, Yu2018, Oishi2019, Onga2019, Washinger2021].
Wachinger et al. \cite{Wachinger2021} analyzed 35,320 MR images from 17 open datasets and performed the 'Name That Dataset' test, that is guessing which dataset it is based on the images alone. They reported a prediction accuracy of 71.5\% based only on volume and thickness information from 70\% of the training data. This is evidence that there are clear features left among datasets.
Removing those variabilities is essential in multi-site and long-term studies and for building a robust CBIR system. There has been an increase in recent research on data harmonization, i.e., eliminating or reducing variation that is not intrinsically related to the brain's biological features.

% Data harmonization（DH）simple-image-based
Perhaps the most straightforward image harmonization approach is to reduce the variations in the intensity profile \cite{Gao2019,Um2019}. In the methods in both \cite{Gao2019} and \cite{Um2019}, correction of the luminance distribution for each sub-region reduces the variability of the underlying statistics between images, whereas histogram equalization reduces the variability of neuroradiological features. However, these methods are limited to approximating rudimentary statistics that can be calculated from images, and they are based on the assumption that  the intensity histogram is similar among images. This assumption is invalid when images that contain pathological findings that affect intensity profile are included. While some improvement in unintended image variability can be expected, the effect on practical tests that utilize data from multiple sites is unknown.

% Data harmonization（DH）statistical batch removal
In the field of genomics, Johnson et al. \cite{Johnson2007} proposed an empirical Bayes-based correction method to reduce batch effects, which are non-biological differences originating from each batch of micro-array experiments obtained from multiple tests. This effective statistical bias reduction method is now called ComBat, and it has recently been published as a tool for MRI harmonization \cite{Fortin2017-COMBAT}. This tool has been applied to several studies \cite{Fortin2017a,Fortin2017b,Yu2018,Wachinger2021}.
%[Fortin2017a, Fortin2017b, Yu2018, Wachinger2021]. 
The ComBat-based methods standardize each cortical region based on an additive  and use multiplicative linear transform to compensate for variability. Some limitations of these models have been pointed out, such as the following: (i) they might be insufficient for complex multi-site and area-level mapping, (ii) the assumption of certain prior probabilities (Gaussian or inverse gamma) is not always appropriate, and (iii) they are susceptible to outliers \cite{Zhao2019}.%\cite{Zhao2019}.

% Data harmonization（DH）deep-learning-based
Recently, advancements in machine learning techniques \cite{Zhao2019,Dewey2019,Moyer2020,Dinsdale2021} 
%[Dewey2019, Zhao2019, Moyer2020, Dinsdale2021] 
have provided practical solutions for MR image harmonization.
DeepHarmony \cite{Dewey2019} uses a fully convolutional U-net to perform the harmonization of scanners.
The researchers used an MRI dataset of multiple sclerosis patients in a longitudinal clinical setting to evaluate the effect of protocol changes on atrophy measures in a clinical study. As a result, DeepHarmony confirmed a significant improvement in the consistency of volume quantification across scanning protocols. This study was practical in that it aimed to directly standardize MR images using deep learning to achieve long-term, multi-institutional quantitative diagnosis.
However, this model requires ''traveling head'' (participants are scanned using multiple MRI scanners) to train the model.
Zhao et al. \cite{Zhao2019} attempted to standardize a group of MR images of infants taken at multiple sites into a reference group using CycleGAN \cite{Zhu2017}, which has a U-net structure in the generator. The experiment validated the evaluation of cortical thickness with several indices (i.e., ROI (region-of-interest)-base, distribution of low-dimensional representations).
They argued that the retention of the patient's age group was superior to ComBat in evaluating group difference.

Moyer et al. \cite{Moyer2020} proposed a sophisticated training technique to reconstruct bias-free MR images by acquiring a low-dimensional representation independent of the scanner and condition. Their method is an hourglass-type unsupervised learning model based on variational autoencoders (VAE) with an  encoder--decoder configuration. 
The input $\vt{x}$ and output $\vt{x'}$ are the same MR images, and their low-dimensional representation is $\vt{z}$ (i.e., $\vt{x}\rightarrow \vt{z} \rightarrow \vt{x'}$). 
The model is trained with the constraint that $\vt{z}$ and site- and scanner-specific information $\vt{s}$ are orthogonal (actually relaxed), such  that the $\vt{s}$ in $\vt{z}$ is eliminated. They demonstrated the advantages of their method on diffusion MRI, but their technological framework is applicable to other modalities.

Dinsdale et al. \cite{Dinsdale2021} also proposed a data harmonization method based on the idea of domain adaptation \cite{Ganin2016}.
Their model uses adversarial learning, where the feature extractor consisting of convolutional neural networks (CNN) following the input is branched into a fully connected net for the original task (e.g., segmentation and classification) and other fully connected nets for domain discriminators (e.g., scanner type or site prediction) to make the domain unknown while improving the accuracy of the original task. They have confirmed its effectiveness in age estimation and segmentation tasks.

% DH for CBIR
The methods developed by Moyer et al. and Dinsdale et al. aim to generate a low-dimensional representation with ''no site information'',  and they are highly practical and generalizable techniques for data harmonization. Nevertheless, for CBIR, a method that is applicable for a large number of legacy images is necessary. Here, it is not realistic to collect images from each site and train the model to harmonize them. Practically, a method that can convert heterogeneous images in terms of variations in scanners and scan parameters into images scanned by a given pseudo-''standard'' environment by applying a learned model is highly desired.

% proposal
In this paper, we propose a novel framework called disease-oriented image embedding with pseudo-scanner  standardization (DI-PSS) to obtain a low-dimensional representation of MR images for practical CBIR implementation. The PSS, the key element of the proposal, corrects the bias caused by different scanning environments and converts the images so that it is as if the same equipment had scanned them. Our experiments on ADNI and PPMI datasets consisting of MR images captured by three manufacturers' MRI systems confirmed that the proposed DI-PSS plays an important role in realizing CBIR.

% highlight
The highlights of this paper's contribution are as follows:
\begin{itemize}
\item To the best of the authors' knowledge, this is the first study of the acquisition and quantitative evaluation of an effective low-dimensional representation of brain MR images for CBIR, including scanner harmonization.
%This is the first study to evaluate the effectiveness of data harmonization in CBIR applications for brain MR images. 
\item Our DI-PSS framework reduces undesirable differences caused by differences in scanning environments (e.g., scanner, protocol, dataset) by converting MR images to images taken on a predefined pseudo-standard scanner, and a deep network using a metric learning acquires a low-dimensional representation that better represents the characteristics of the disease.
%We propose DI-PSS to acquire a low-dimensional representation that is less sensitive to the environment and the individual and more sensitive to the disease-related morphological characteristics, which is desirable for the realization of CBIR.
\item DI-PSS provides appropriately good low-dimensional representations for images from other vendors’ scanners, diseases, and datasets that are not used for learning image harmonization. This is an important feature for the practical and robust CBIR, which applies to a large amount of legacy MRIs scanned at heterogeneous environments.
%We demonstrate the potential of our DI-PSS for harmonizing images scanned by different scanners or using a scan protocol that was not used to scan the training data. This is an important feature for the practical and robust CBIR, which applies to a large amount of legacy MRIs scanned at heterogeneous environments.
\end{itemize}

% section 2
\section{Clarification of the issues addressed in this paper}
\label{sec:sec2}
% 2.1
\subsection{Overlooking the problem}
We begin by presenting the issues to be solved in this paper.
As mentioned above, to realize CBIR for brain MRI, Onga et al. proposed a new technique called disease-oriented data concentration with metric learning (DDCML), which acquires low-dimensional representations of 3D brain MR images that are focused on disease features rather than the features of the subject's brain shape \cite{Onga2019}.
DDCML is composed of 3D convolutional autoencoders (3D-CAE) effectively combined with deep metric learning. Thanks to its metric learning, DDCML could acquire reasonable low-dimensional representations for unlearned diseases according to their severity, demonstrating the feasibility of CBIR for brain MR images. However, we found that such representations are highly sensitive to differences in datasets (i.e., differences in imaging environments, scanners, protocols, etc.), which is a serious challenge for CBIR.

\begin{figure}[t]
%\centering
\includegraphics[width=0.9\linewidth]{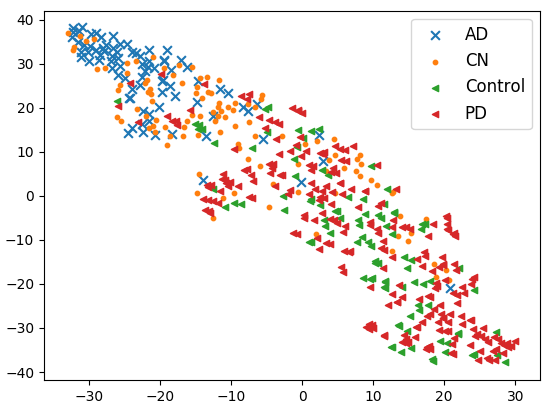}
\caption{Plots of low-dimensional representations of 3D MRI obtained from different datasets.}
\small{The impact of different scanners (CN$\Leftrightarrow$Control; they are medically equivalent) is greater than the impact of the disease (AD$\Leftrightarrow$CN).}
\label{fig:fig1}
\end{figure}

Figure \ref{fig:fig1} shows the low-dimensional distribution obtained by DDCML and visualized by t-SNE \cite{Laurens2008}. Here, DDCML was trained on Alzheimer's disease (AD) and healthy cases (clinically normal; CN) in the ADNI2 dataset and evaluated ADNI2 cases not used for training and healthy control (Control -- equivalent to CN) and Parkinson's disease (PD) cases in the untrained PPMI dataset. From the perspective of CBIR, it is desirable to obtain similar low-dimensional representations for CN and Control. However, it can be confirmed that the obtained low-dimensional representations are more affected by the differences in the environment (dataset) than by the disease. As mentioned above, differences in imaging environments, including scanners, are a major problem in multi-center and time series analysis, and inconsistent low-dimensional representations because of such differences in datasets are a fatal problem in CBIR implementation. The purpose of this paper is reducing these differences and to obtain a low-dimensional representation that better captures the characteristics of the disease and is suitable for appropriate CBIR.

%
% Insert Fig. 1
%

\subsection{Our data harmonization strategy for realizing CBIR}
In studies dealing with multi-site and long-term data, it is undoubtedly important to reduce non-biological bias originating from differences among sites and datasets.
Since the methods of Moyer et. al.\cite{Moyer2020} and Dinsdale et al. \cite{Dinsdale2021} are theoretical and straightforward learning method that utilizes images of the target site to achieve data harmonization, their robustness to unexpected input (i.e. from another site or dataset) is questionable. Therefore, in principle, the images of all target sites (scanners, protocols) need to be learned in advance.
Since CBIR requires more consideration of the use of images taken in the past, the number of environments that need to be addressed can be larger than for general data harmonization.
It will be more difficult to implement a harmonization method that learns all the data of multiple environments in advance.
Therefore, in contrast to their approaches, we aim to achieve data harmonization by converting images taken in each environment into images that can be regarded as having been taken in one predetermined ''standard'' environment (e.g., the scanner currently used primarily at each site). However, in addition to the problems described above, it is practically impossible to build an image converter for each environment.

With this background, we have developed a framework that combines CycleGAN, which realizes robust image transformation, with deep metric learning to achieve a certain degree of harmonization even for images in untrained environments. In this paper, we validate the feasibility of our framework, which converts MR images captured in various environments into pseudo standard environment images using only one type of image converter.

%% section 3
\section{Disease-oriented image embedding with pseudo-scanner  standardization (DI-PSS)}
\label{sec:sec3}
The aim of this study is to obtain a low-dimensional embedding of brain MRI that is independent of the MRI scanner and individual characteristics but dependent on the pathological features of the brain, to realize a practical CBIR system for brain MRI.
To accomplish this, we propose a DI-PSS framework, which is composed of the three following components: (1) pre-process, (2) PSS, and (3) embedding acquisition.

%3.1
\subsection{The Pre-processing component (skull stripping with geometry and intensity normalization)}
The pre-processing component performs the necessary pre-processing for future image scanner standardization processing and low-dimensional embedding acquisition processing. 
Specifically, for all 3D brain MR image data, skull stripping was performed using a multi-atlas label-fusion algorithm implemented in the MRICloud \cite{Mori2016}.
The skull-stripped images were linearly aligned to the JHU-MNI space using a 12-parameters affine transformation function implemented in the MRICloud, resulting in aligned brain images. 
This feature makes a significant contribution to the realization of the proposed PSS in the next stage. It is important to note here that since brain volume information is the feature that contributes most to the prediction of the dataset \cite{Wachinger2021}, the alignment to a standard brain with this skull stripping technique should also contribute to the harmonization of the data. 
In addition, because the intensity and contrast of brain MR images are arbitrarily determined, there is a large inter-image variation. In brain MR image processing using machine learning, the variation in the average intensity confounds the results. Therefore, we standardized the intensity so that the average intensity value of each case was within mean $\mu=18$ and margin $\epsilon = 1.0$ by performing an iterative gamma correction process, as in previous studies \cite{Arai2018}\cite{Onga2019}.
%[Arai2018, Onga2019].

%3.2
\subsection{The PSS component}
% 3.2.1
\subsubsection{The concept of PSS}
The proposed PSS is an image conversion scheme that converts a given raw MR image into a synthesized image that looks like an MR image scanned by a standard scanner and a protocol. Since there are numerous combinations of scanners and scan parameters, building scanner- and parameter-specific converters is not practical. 
Therefore, in our PSS scheme, we only construct a 1:1 image conversion model (i.e., PSS network) that converts images from a particular scanner $Y$ to a standard scanner $X$.
That is, a particular PSS network is used to convert images captured by other scanners $(Z_1, Z_2,\cdots)$ as well. This strategy is in anticipation of the generalizability of the PSS network, backed by advanced deep learning techniques.
In this paper, we evaluate the robustness of our image transformations provided by PSS on MR images taken by other vendors' scanners and on images in different datasets.

\begin{figure*}[t]
\centering
\includegraphics[width=0.9\linewidth]{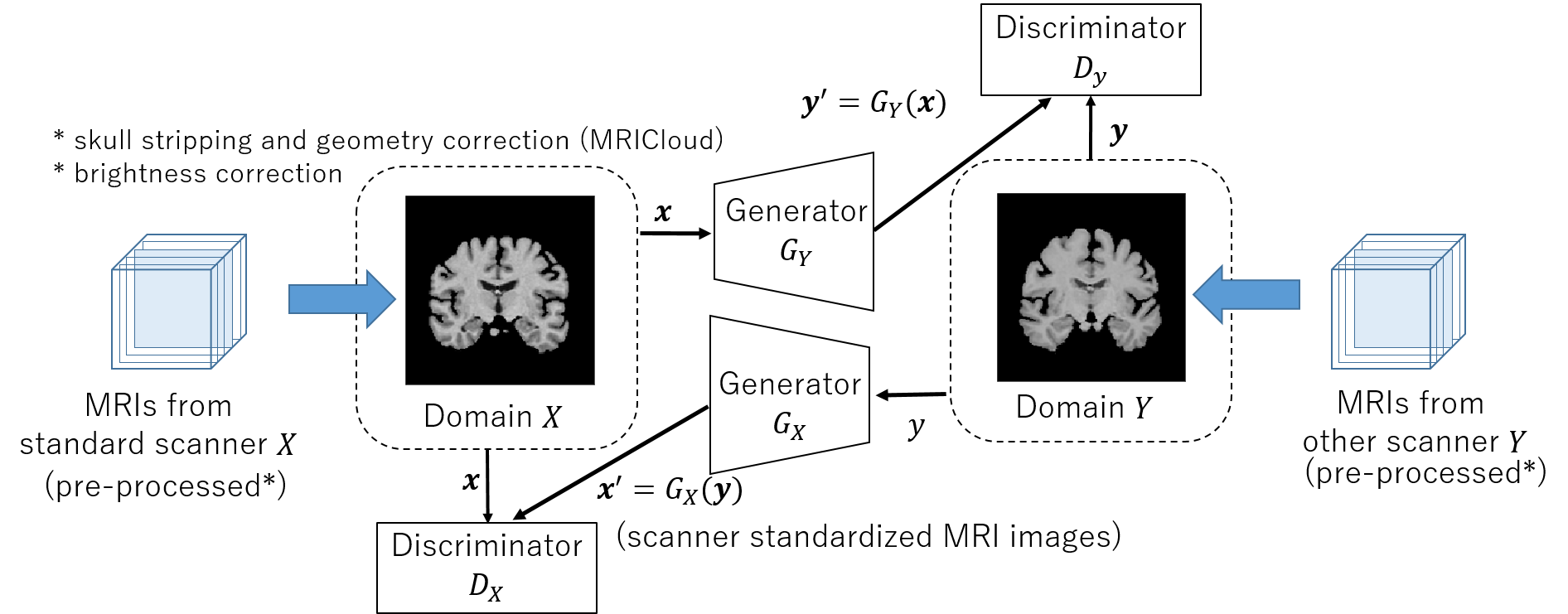}
\caption{Overview of pseudo-scanner  standardization (PSS) network.}
\small{Our PSS network is based on CycleGAN, and PSS is performed with trained generator $G_X$.}
\label{fig:fig2}
\end{figure*}

Figure \ref{fig:fig2} gives an overview of our PSS network that realizes the PSS. The PSS network makes effective use of CycleGAN \cite{Zhu2017}, which has achieved excellent results in 1:1 image transformation. 
Here, training of CycleGAN generally requires a lot of training data, especially in the case of 3D data, because the degree of freedom of the model parameters is large. However, it is difficult to collect such a large amount of supervised labeled 3D MRI data to keep up with the increase. Since the position of any given slice is almost the same in our setting thanks to MRICloud in the skull stripping process, a 3D image can be treated as a set of 2D images containing position information. With these advantages, our PSS suppressed the problems an  overwhelmingly insufficient amount of training data and the high degree of freedom of the transformation network. In sum, arbitrary slices are cut out from the input 3D image and converted to slices corresponding to the same position in the 3D image as the target domain using the PSS network based on common (2D) CycleGAN. Note that the PSS process is performed using the trained generator $G_X$.

%
% Insert Fig 2
%

% 3.2.2
\subsubsection{Implementation of the PSS network}
The structure of the PSS network that realizes the proposed PSS is explained according to the CycleGAN syntax, with images captured by a standard scanner as domain $X$ and images captured by a certain different scanner as domain $Y$.
Generator $G_Y$ transforms (generates) an image $\vt{y’}=G_Y(\vt{x})$ with the features of domain $Y$ from an image $\vt{x}$ of the original domain $X$.
Discriminator $D_Y$ determines the authenticity of the real image $\vt{y}$ belonging to domain $Y$ or the generated $\vt{y'}=G_Y(\vt{x})$.
Similarly, the conversion from domain $Y$ to domain $X$ is performed by generator $G_Y$, and discriminator $D_X$ judges the authenticity of the image. The goal of this model is to learn maps of two domains $X$ and $Y$ given as training data. Note here again that we use the trained module $G_X$ (maps $Y$ to $X$) as an image converter.

The training of the model proceeds by repeating the transformation of the training data sample $\vt{x_i} \in X$ and the training data sample $\vt{y_j} \in Y$. %(j=1,2,\cdots,N_Y)$.
The overall objective function of the PSS network, $L_{PSS}$ to be minimized, consists of the three following loss components: adversarial loss ($L_{GAN}$), cycle consistency loss ($L_{eye}$), and identity mapping loss ($L_{identity}$). This is expressed as follows:
\begin{equation}
\begin{split}
    L_{PSS} (G_Y,&G_X,D_Y,D_X) = \\
    &L_{GAN} (G_Y,D_Y) + L_{GAN} (G_X,D_X) \\ 
    \hspace*{3em}&+\lambda_1 L_{eye}+ \lambda_2 L_{identity}.
\end{split}
\end{equation}
The adversarial loss ($L_{GAN}$) is defined based on the competition between the generator, which tries to produce the desired other domain image, and the discriminator, which sees through the fake generated image; this minimization implies a refinement of both. From the point of view of image transformation, the minimization of this loss means that the probability distribution generated by the generator is closer to the probability distribution of the counterpart domain, which means that a higher quality image can be obtained. This loss is defined in both directions, $X\rightarrow Y$ and $Y\rightarrow X$, and these are expressed in order as follows:
\begin{equation}
\begin{split}
L_{GAN}(G_Y,D_Y) = 
 &E_{\vt{y}\sim p_{data}(\vt{y})}[(D_Y(\vt{y})-1)^2] \\
   +&E_{\vt{x}\sim p_{data}(\vt{x})}[(D_Y(G_Y(\vt{x}))^2], 
%   &E_{\vt{y}\sim p_{data} (\vt{y})} \log D_Y (\vt{y}) \\
%  +&E_{\vt{x}\sim p_{data}(\vt{x})} \log ( 1-D_Y(G_Y (\vt{x}))
\end{split}
\end{equation}
\begin{equation}
\begin{split}
L_{GAN}(G_X,D_X) = 
    &E_{\vt{x}\sim p_{data}(\vt{x})}[(D_X(\vt{x})-1)^2] \\
   +&E_{\vt{y}\sim p_{data}(\vt{y})}[(D_X(G_X(\vt{y}))^2]. 
  %  &E_{\vt{x}\sim p_{data}(\vt{x})} \log D_X (\vt{x}) \\
  % +&E_{\vt{y}\sim p_{data}(\vt{y})} \log ( 1-D_X(G_X(\vt{y}))
\end{split}
\end{equation}
The cycle consistency loss ($L_{eye}$) is a constraint to guarantee that mutual transformation is possible by cycling two generators:
\begin{equation}
\begin{split}
L_{eye}(G_X,G_Y) = 
 &E_{\vt{x} \sim p_{data} (\vt{x})} || G_X (G_Y (\vt{x}))-\vt{x}||_1 \\
+ &E_{\vt{y} \sim p_{data} (\vt{y})} ||G_Y (G_X (\vt{y}))-\vt{y}||_1
\end{split}
\end{equation}
Finally, the identity mapping loss ($L_{identity}$) is a constraint to maintain the original image features without performing any transformation when the image of the destination domain is input:
\begin{equation}
\begin{split}
L_{identity}(G_X,G_Y)=
 &E_{\vt{x} \sim p_{data} (\vt{x})} || G_X(\vt{x})-\vt{x}||_1 \\
+&E_{\vt{y} \sim p_{data} (\vt{y})} ||G_Y(\vt{y})-\vt{y}||_1
\end{split}
\end{equation}
It has been confirmed that the introduction of this constraint can suppress the learning of features that are not important in either domain, such as unneeded tints. Here, $\lambda_1$ and $\lambda_2$ are hyper-parameters and we set $\lambda_1 =10.0$ and $\lambda_2=0.5$ as in the original setting.

%3.3
\subsubsection{The Embedding acquisition component}
In the embedding acquisition component, the low-dimensional embedding of 3D  brain MRI  images is obtained by our embedding network after the PSS process.
Our embedding network is a 3D-CAE model consisting of encoders and decoders with distance metric learning, referring to Onga et al.'s  DDCML\cite{Onga2019}.

Distance metric learning is a learning technique that reduces the Euclidean distance between feature representations of the same label and increases the distance between feature representations of different labels. 
Thanks to the introduction of metric learning, 3D-CAE has been found to yield embedding that is more focused on disease features.
%
%
%The distance distribution in the low-dimensional embedding space for input $\vt{x}$ and the data $\vt{x_i}$ ($i \in 1, \cdots c$; where $c$ is the number of types of disease labels in the dataset) in the low-dimensional feature space is calculated by the following equation using the criterion of Hoffer et al. \cite{Hoffer2016},
%
%\begin{equation}
%\begin{split}
%P(\vt{x};\vt{x_1},\cdots, \vt{x_c} )_i &= \frac{ \exp(-{||f(\vt{x})-f(\vt{x_i})||}^2 }{ %\sum_j^c \exp(-{||f(\vt{x})-f(\vt{x_j})||}^2)}\\
%& i \in \{1,\cdots, c\}
%\end{split}
%\end{equation}
%
%,where $(\vt{x_1},\cdots,\vt{x_c})$ is randomly selected data from each class $i$.
%$P(\vt{x};\vt{x_1},\cdots,\vt{x_c})_i$ denotes the probability that the data $\vt{x}$ is classified into each class $i$, and $f$ denotes the operation by encoder (i.e., encoder part of the 3D-CAE in our implementation).
%

According to Hoffer’s criteria \cite{Hoffer2016}, the distance distribution in the low-dimensional embedding space for input $\vt{x}$ for class $i$ ($i \in 1, \cdots c$; where $c$ is the number of types of disease labels in the dataset) is calculated by
\begin{equation}
P(\vt{x};\vt{x_1},\cdots, \vt{x_c} )_i = \frac{ \exp(-{||f(\vt{x})-f(\vt{x_i})||}^2) }{ \sum_j^c \exp(-{||f(\vt{x})-f(\vt{x_j})||}^2)}.
\end{equation}
Here, $\vt{x_i}$ ($i \in 1, \cdots, c$) is randomly sampled data from each class $i$, and $f$ denotes the operation of the encoder (i.e., encoder part of the 3D-CAE in our implementation). This probability can be thought of as the probability that the data $\vt{x}$ belong to each class $i$.

The loss function $L_{dist}$ is calculated by the cross-entropy between the $c$-dimensional vector $\vt{P}$ described above and the $c$-dimensional one-hot vector $\vt{I}(\vt{x})$ with bits of the class to which $\vt{x}$ belongs as 
\begin{equation}
L_{dist} (\vt{x},\vt{x_1},\cdots,\vt{x_c}) = H(\vt{I}(\vt{x}),\vt{P}(\vt{x};\vt{x_1},\cdots,\vt{x_c}))
\end{equation}
Here, $H(\vt{I}(\vt{x}),\vt{P}(\vt{x};\vt{x_1},\cdots,\vt{x_c}))$ takes a small value when the probability that the element firing in $\vt{I}(\vt{x})$ belongs to the class it represents is high, whereas it takes a large value when the probability is low. Thus, $L_{dist}$ aims at the distribution of the sampled data at locations closer to the same class and farther from the different classes on the low-dimensional feature space.
Finally, the objective function $L_{CAE}$ of our low-dimensional embedding acquisition network consisting of 3D-CAE and metric learning is finally expressed by the following equation:
\begin{equation}
L_{CAE}=L_{RMSE} + \alpha L_{dist} (\vt{x},\vt{x_1},\cdots,\vt{x_c} )
\end{equation}
Here, $L_{RMSE}$ is the pixel-wise root mean square error normalized by image size in CAE image reconstruction. Furthermore, $\alpha$ is a hyper-parameter set to 1/3 based on the results of preliminary experiments. 

%%section 4
\section{Experiments}
\label{sec:sec4}
In CBIR, cases of the same disease should be able to acquire similar low-dimensional representations, regardless of the individual, scanner, or protocol. We investigated the effectiveness of the proposed DI-PSS by quantitatively evaluating how PSS changes the distribution of embeddings within and between data groups (i.e., combination of scanner type and disease). In addition, we compared the clustering performance of the obtained embeddings against diseases with and without PSS.

%4.1
\subsection{Dataset}
In this experiment, we used the ADNI2 and PPMI datasets, in which the vendor information of the scanners (Siemens [SI], GE Medical Systems [GE], Philips Medical Systems [PH]) was recorded along with the disease information. Statistics of those datasets used in the experiment are shown in Table \ref{table:table1}.
We used Alzheimer's disease (ADNI-AD or AD) and clinically normal cases (ADNI-CN) from ADNI2 dataset with vendor information.
From the PPMI dataset, we used two types of labeled images, Parkinson's disease (PD) and Control. We did not utilize the scanner information for this dataset in evaluating the versatility of the proposed method. 
Note that ADNI-CN and Control can be considered medically equivalent. Furthermore, PD is known to show little or no difference in MRI from healthy cases \cite{Postuma2015}\cite{Meijer2017}.
%[Postuma2015, Meijer2017].

The ADNI and PPMI are longitudinal studies that include multiple time points, and the datasets contain multiple scans for each participant. To avoid duplication, one MRI was randomly selected from each participant. 
The MRICloud (\url{https://mricloud.org/}) was used to skull strip the T1-weighted MRIs and affine transform to the JHU-MNI atlas \cite{Oishi2009}. %[Oishi, 2009].
A neurologist with more than 20 years of experience in brain MRI research performed the quality control of the MRIs and removed MRIs that the MRICloud did not appropriately pre-process.
Due to the neural network model used in the experiments, the skull-stripped and affine-transformed brain MR images were converted to 160$\times$160$\times$192 pixels after cropping the background area.
Training and evaluation of the PSS network and embedding network were performed using five-fold cross validation. In the evaluation experiments described below,, the evaluation data of each fold is not included in the training data for either the PSS network or the  embedding network.
Note that even skilled and experienced neuroradiologists cannot separate PD from CN or Control by visual inspection of the T1-weighted images.
Therefore, we did not expect these two conditions to be separable by unsupervised clustering methods even after applying the DI-PSS.

%
% Table 1
%
\begin{table}[t]
%\centering
\caption{Dataset used in our study}
\label{table:table1}
\begin{tabular}{ccllccc}
\toprule
\multicolumn{2}{c}{dataset}   & vendor & label & \#used & \#patients & \#total \\ \hline
\multirow{6}{*}{ADNI} & \multirow{3}{*}{CN} & Siemens & CN\_SI & 92 & 103 & 439 \\
 &  & GE & CN\_GE & 93 & 101 & 494 \\
 &  & Philips & CN\_PH & 27 & 27 & 119 \\ \cline{2-7}
 & \multirow{3}{*}{AD} & Siemens & AD\_SI & 80 & 84 & 254 \\
 &  & GE & AD\_GE & 80 & 92 & 302 \\
 &  & Philips & AD\_PH & 20 & 24 & 73 \\ \hline
\multirow{2}{*}{PPMI} &  & \multirow{2}{*}{n/a} & Control & 75 & 75 & 114 \\
 &  &  & PD & 149 & 149 & 338 \\
 \bottomrule
 \end{tabular}
\footnotesize{ADNI-CN and Control can be considered medically equivalent. \\
 There are no PD-related anatomical features observable on T1-weighted MRI.}
\end{table}

%
% Fig 3 (PSS network architecture)
\begin{figure*}[t]
\begin{minipage}[t]{0.57\hsize}
\centering
\includegraphics[width=0.95\linewidth]{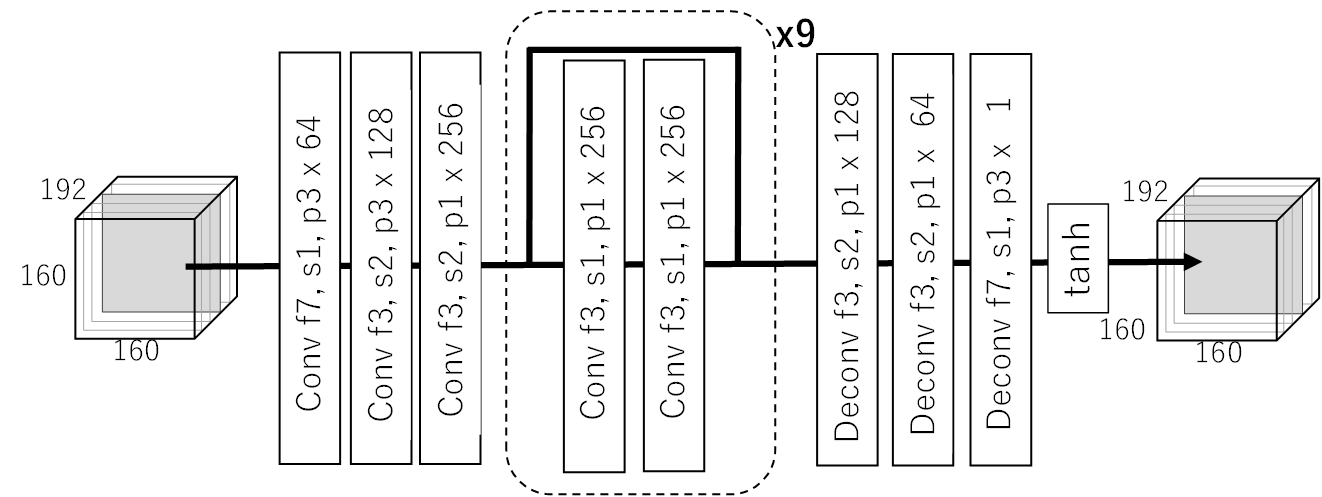}
\subcaption{Generator}
\label{fig3a}
\end{minipage}
\begin{minipage}[t]{0.38\hsize}
\centering
\includegraphics[width=0.95\linewidth]{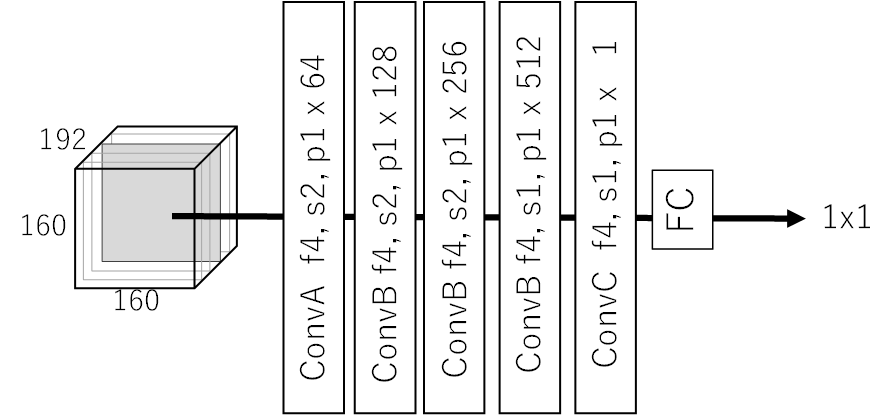}
\subcaption{Discriminator}
\label{fig3b}
\end{minipage}
\caption{Architecture of (a) Generators ($G_X, G_Y$) and (b) Discriminators ($D_X, D_Y$) in the PSS network.}
\footnotesize{(a) kernel size (f$\times$f), stride size (s), padding size (p),  $\times$ \# of kernel + instance norm + ReLU*\\
(b) convA : kernel size (f$\times$f), stride size (s), padding size (p),  $\times$ \# of kernel + LeakyReLU\\
\hspace*{1em} convB : kernel size (f$\times$f), stride size (s), padding size (p),  $\times$ \# of kernel + instanceNorm + LeakyReLU\\
\hspace*{1em} convC : kernel size (f$\times$f), stride size (s), padding size (p),  $\times$ \# of kernel  \\
FC : fully connected layer (400$\rightarrow$1) }
\label{fig:fig3}
\end{figure*}
%

%
% Fig 4 (3D CAE)
%
\begin{figure*}[t]
\centering
\includegraphics[width=0.95\linewidth]{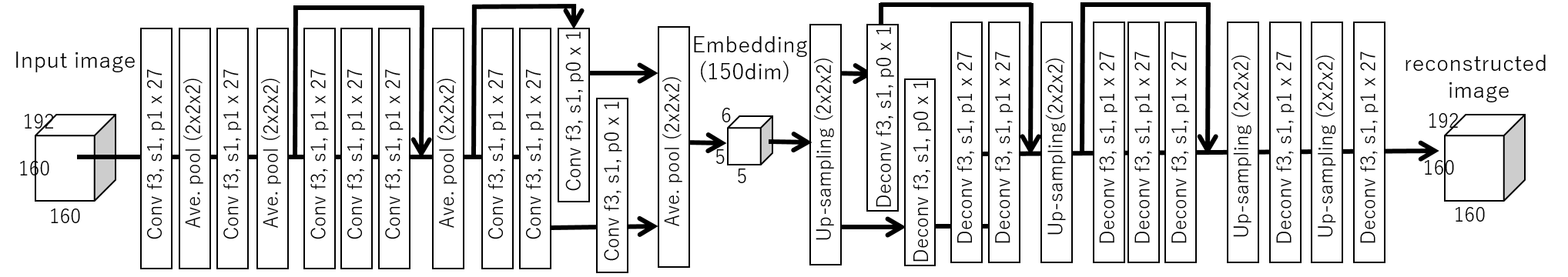}
\caption{Architecture of embedding network.}
\footnotesize{conv : kernel 3$\times$3, stride size=1, padding size=1,  $\times$  (\# of kernel) + ReLU\\
deconv : kernel 3$\times$3, stride size=1, padding size=1,  $\times$  (\# of kernel) + ReLU\\
average pooling, up-sampling (bi-linear interpolation): 2$\times$2$\times$2:1.}
\label{fig:fig4}
\end{figure*}

%4.2 
\subsection{Detail of the PSS network and its training}
Figures \ref{fig:fig3}a and \ref{fig:fig3}b show the architecture of the generator ($G_X, G_Y$) and the discriminator ($D_X, D_Y$), respectively of the PSS network. 
They are basically the same as the original CycleGAN for 2D images. 
Since PSS is to reduce the bias caused by variations in scanners and scan parameters, the disease-related anatomical variations should be minimized in the training images.
Therefore, we used only ADNI-CN cases, in which disease features do not appear in the brain structure, to train the PSS network. 
In this experiment, we chose the Siemens scanner as the standard scanner because it has largest market share, and we chose the GE scanner as the specific vendor of image conversion source.
In other words, our PSS network is designed to convert CN images taken by GE scanners from the ADNI2-dataset (CN\_GE) to synthetic images similar to those scanned by the Siemens scanners (CN\_SI). We evaluated the applicability of the PSS to the diseased brain MRIs (AD and PD), as well as the generalizability to the non-GE scanners (see Section IV.D).  In PSS network, we used coronal images for the training. The number of training images of each fold in the PSS network is (93+92)$\times$4/5 (5-fold CV)$\times$192 (slices).

%4.3
\subsection{Detail of the embedding network and its training}
Figure \ref{fig:fig4} shows the architecture of our 3D-CAE-based embedding network.
Our embedding network embeds each 3D brain MR image into 150-dimensional vectors. The size of the MRIs handled by the embedding network is halved at each side, as in DDCML \cite{Onga2019}, to improve the learning efficiency. Note that the compression ratio of our embedding network is (80$\times$80$\times$96):150 = 4,096:1. The embedding network was trained and evaluated using ADNI2 and PPMI datasets with the five-fold cross-validation strategy.

As mentioned above, PD and CN cannot even be diagnosed from images by skilled neuroradiologists, so for training 3D-CAE to obtain low-dimensional representations, two classes of metric learning are used so that the representations of AD and (CN + Control) are separated. The low-dimensional representations of brain MR images are acquired by five-fold cross validation of 3D-CAE. In addition to AD, CN, and Control in each test fold, the low-dimensional representation of PD, which was not included in the training, is analyzed to quantitatively verify the effectiveness of the proposed DI-PSS evaluation.

%4.4
\subsection{Evaluation of the PSS}
To evaluate the effectiveness of the proposed DI-PSS framework, we evaluate the three following elements: 
\begin{enumerate}
\item Changes in MR images
\item Distribution of the embedding.
\item Clustering performance of the embedding.
\end{enumerate}

In (1), we assess how the images are changed by our scanner standardization.
We quantitatively evaluate the difference between the original (raw) image and the synthetic image with peak signal-to-noise ratio (PSNR),  root mean squared error (RMSE), and structured similarity (SSIM). 
To ensure that the evaluation is not affected by differences in brain size, these evaluations were performed on brain regions only.
Although MRICloud, which is used in skull stripping in this experiment, standardizes the brain size to the standard brain size, reducing the differences in brain size between cases, this method was adopted for a more rigorous evaluation.

In (2), we quantitatively examine the effect of PSS by analyzing the distribution of the obtained low-dimensional representations.
Specifically, for each category (e.g., CN\_SI, AD\_GE) we investigate the following: (i) variation (i.e., standard deviation) of the embedding and (ii) the mean and standard deviation of the distance from each embedding to the centroid of a different category, where the distance between the centroids of ADNI-CN\_Siemens (CN\_SI) and ADNI-AD\_Siemens (AD\_SI) are normalized to 1. 
In addition, we visualize those distributions in 2D space using t-SNE \cite{Laurens2008} as supplemental results for intuitive understanding.

In (3), we evaluate the separability of the resulting embeddings. 
In this study, we performed spectral clustering \cite{Ng2001} to assess its potential quality for CBIR.
In the spectral clustering, we used a normalized graph Laplacian based on 10-nearest neighbor graphs with a general Gaussian-type similarity measure. 
We set the number of clusters to be two (AD vs. CN + Control + PD), which is the number of disease categories to be classified. Here, the consistency of the distance between the embedded data because of the difference in folds is solved by standardizing the distance between CN\_SI and AD\_SI per fold to be 1, as mentioned above.

The clustering performance was evaluated using two methodologies.
The first was evaluation with six commonly used criteria (i.e., silhouette score, homogeneity, completeness, V-measure, adjusted Rand-index [ARI], and adjusted mutual information [AMI]) implemented on  the scikit-learn machine learning library (\url{https://scikit-learn.org/}).
The other is a diagnostic capability based on clustering results.
Here, as with other clustering evaluations in the literature, we swap the columns so that each fold results in the optimal clustering result  and then sum them.

%% section 5
\section{Results}
\label{sec:sec5}
%5.1
\subsection{Changes in MR images by PSS}
\begin{figure*}[t]
\centering
\includegraphics[width=0.9\linewidth]{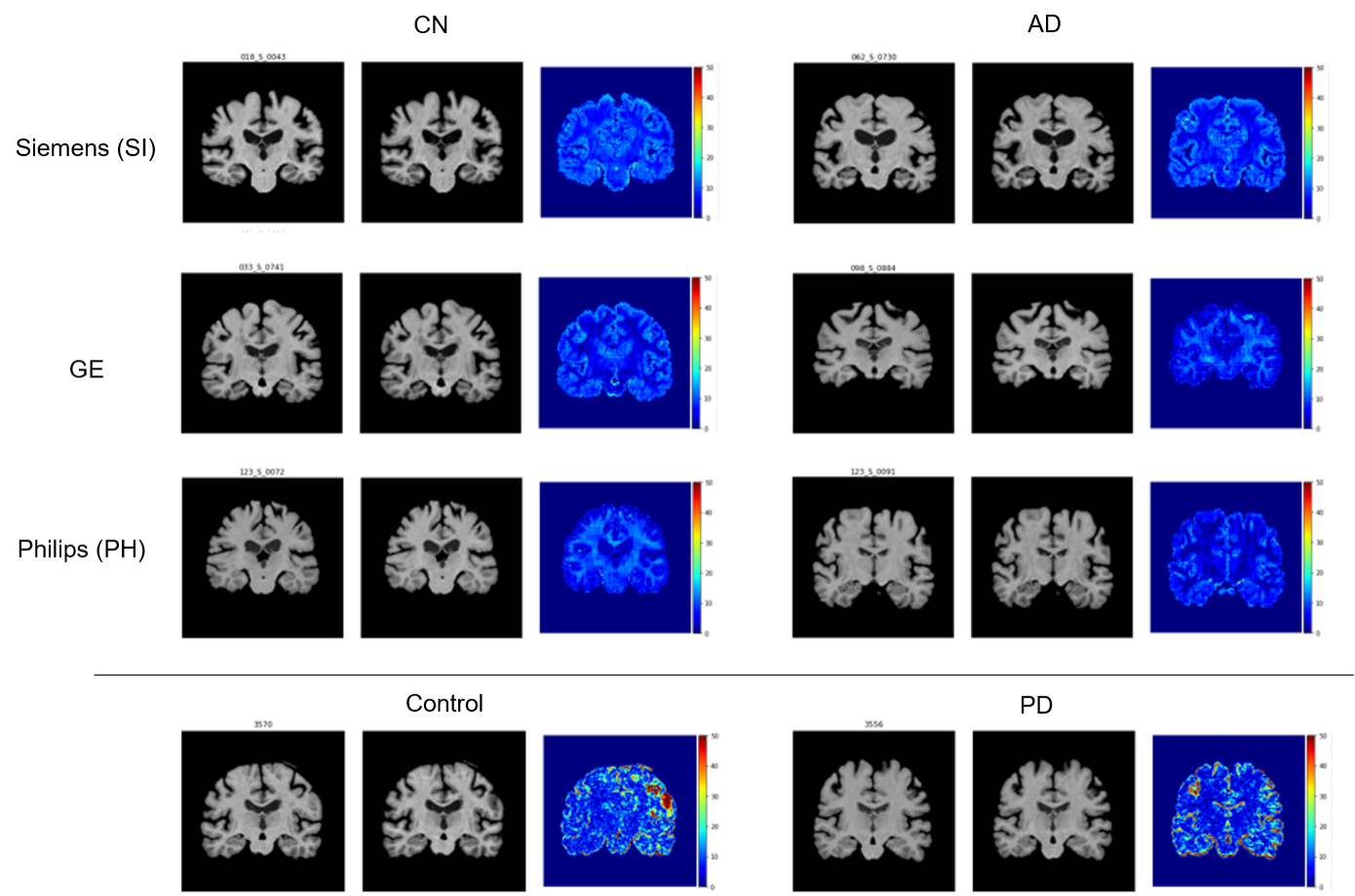}
\caption{Example of image change by PSS (in coronal plane).}
\footnotesize{In each category, from left to right, the original image, the PSS processed image, and the difference between them.}
\label{fig:fig5}
\end{figure*}

Figure \ref{fig:fig5} shows an example of each MR image converted to an image taken on a pseudo-standard (= Siemens) scanner with PSS and the difference visualized.
Table \ref{table:table2} summarizes the statistics of the degree of change in the images in the brain regions. Here, the background region was excluded from the calculation to eliminate the effect of differences in brain size. 
For the ADNI dataset, the differences obtained by the PSS image transformation were not significant between CN, AD, and scanner vendors, although the Philips scanners showed less variation on average.
For the PPMI dataset that was not used for training, the change in the image because of PSS is clearly larger compared with ADNI (approx. $\times$ 1.5 in RMSE). 
In all categories, the amount of change because of PSS varied from case to case, but the PSS treatment did not cause any visually unnatural changes in the images. 

\begin{figure*}[t]
\centering
\begin{minipage}[t]{0.45\hsize}
\centering
\includegraphics[width=0.95\linewidth]{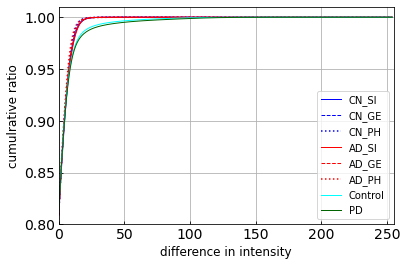}
\subcaption{overall view}
\label{fig6a}
\end{minipage}
\begin{minipage}[t]{0.45\hsize}
\centering
\includegraphics[width=0.95\linewidth]{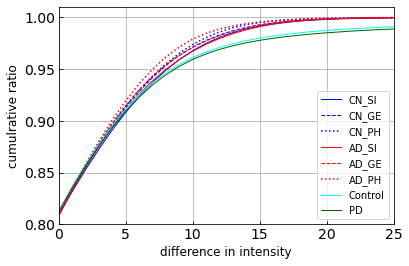}
\subcaption{enlarged view}
\label{fig6b}
\end{minipage}
\caption{Cumulative intensity changes of MR images by PSS.}
\label{fig:fig6}
\end{figure*}

Figure \ref{fig:fig6} shows the cumulative intensity changes of images by PSS in each category. This time, the background areas other than the brain are also included in the evaluation. The number of pixels where the intensity has not changed because of PSS exceeds 80\% for all categories, indicating that no undesired intensity changes have occurred for the background (as also seen in Figures. \ref{fig:fig5} and \ref{fig:fig6}). 
There is no significant difference in the distribution of intensity change by vendor, and the PPMI dataset has a larger amount of intensity change overall.

%
% Fig 5 (PSS image),  
%

%
%Fig 6 (cumulative curve)
%

%
%Table 2 (change statistics)
%
\begin{table}[t]
\centering
\caption{Summary of image changes by PSS}
\label{table:table2}
\begin{tabular}{clccc}
\toprule
dataset  & label & PSNR (db) & RMSE & SSIM \\ \hline
\multirow{6}{*}{ADNI} 
 &  CN\_SI & 31.52 ± 2.85 & 7.17 ± 2.57 & 0.9743 ± 0.0048\\
 &  CN\_GE & 31.67  ±  2.45 & 6.94 ± 2.13 & 0.9748 ± 0.0041 \\
 &  CN\_PH & 32.18 ± 2.65 & 6.58 ± 2.02 & 0.9747 ± 0.0038 \\
 &  AD\_SI & 31.64 ± 3.04 & 7.13 ± 2.77 & 0.9746 ± 0.0043 \\
 &  AD\_GE & 31.65 ± 2.52 & 6.98 ± 2.32 & 0.9750 ± 0.0044 \\
 &  AD\_PH & 32.33 ± 2.13 & 6.36 ± 1.63 & 0.9751 ± 0.0031 \\ \hline
\multirow{2}{*}{PPMI} &  Control & 30.16 ± 5.47 & 9.81 ± 8.32 & 0.9596 ± 0.0346 \\
   & PD & 29.40 ± 5.94 & 11.60 ± 10.89 & 0.9539 ± 0.0473 \\
 \bottomrule
 \end{tabular}
\end{table}

% 5.2
\subsection{Distribution of low-dimensional embedded data}
%5.2.1
\subsubsection{Distance between centers of the data distribution by category}
Table \ref{table:table3} shows the variation (standard deviation; SD) of the 150-dimensional embedded representation in each category. Again, it should be noted here that CN\_SI and AD\_SI were normalized to 1.
The average reduction in SD for all data by PSS was 8.27\%.
Tables \ref{table:table4} shows the statistics of distances from each embedding to the centroid of a different category. This shows the distribution of the data, considering the direction of variation, which is more practical for CBIR application.
With PSS, the average distance between centroids across categories is almost unchanged, but the variability is greatly reduced for all categories.

%
% Table 3
%
\begin{table}[t]
%\centering
\caption{Variation (SD) of the embedding in category $\dagger$}
\label{table:table3}
\begin{tabular}{clcccc}
\toprule
dataset  & label & \#data & baseline & with PSS & $-$ SD (\%)\\ \hline
\multirow{6}{*}{ADNI} 
 &  CN\_SI & 92 & 0.697 &	0.648 &	7.12 \\
 &  CN\_GE & 93 & 0.784 &	0.716 &	8.66 \\ 
 &  CN\_PH & 27 & 0.622 &	0.619 &	0.48 \\
 & ADNI-CN &212 & 0.753 &	0.701 &	6.91 \\ \cline{2-6}
 &  AD\_SI & 80 & 0.863 &	0.783 &	9.24 \\
 &  AD\_GE & 80 & 0.849 &	0.806 &	5.09   \\
 &  AD\_PH & 20 & 0.771 & 	0.706 &	8.46   \\ 
 & ADNI-AD &180 & 0.876 &   0.823 & 6.09   \\ \hline
\multirow{2}{*}{PPMI} 
 &Control & 75 & 0.607 & 0.554 & 8.74   \\ 
 &PD      & 149& 0.603 & 0.515 & 14.71   \\ \hline
 both & CN & 92 & 0.759 & 0.704 & 7.20 \\ \hline
 \multicolumn{2}{c}{all} &616& 0.755& 0.693 & 8.27 \\
 \bottomrule
 \end{tabular}
 \small{$\dagger$:Distance between CN\_SI and AD\_SI were normalized to 1.}
\end{table}

%
% Table 4
%
\begin{table}[t]
%\centering
\caption{Mean and variability of embedding across categories of data $\dagger$}
\label{table:table4}
\begin{tabular}{llrrrrr}
\toprule
\multicolumn{2}{c}{}&\multicolumn{2}{c}{baseline} & \multicolumn{2}{c}{with PSS} & -SD (\%) \\
from & to & mean & SD & mean & SD &   \\ \hline
ADNI-CN & AD & \multirow{2}{*}{0.879} & 0.669  & \multirow{2}{*}{0.890}& 0.541 & 19.1 \\ 
AD & ADNI-CN &                        & 0.875  &  & 0.729  & 16.6  \\
Control & AD & \multirow{2}{*}{1.354} & 0.745  & \multirow{2}{*}{1.329}& 0.537  & 28.0 \\
AD & Control &                        & 0.907  &                       & 0.702  & 22.6  \\
PD & Control & \multirow{2}{*}{0.256} & 0.469  & \multirow{2}{*}{0.297}& 0.312  & 33.5  \\
Control & PD &                        & 0.414  &                       & 0.368  & 11.2  \\
ADNI-CN & PD & \multirow{2}{*}{0.364} & 0.609  & \multirow{2}{*}{0.362}& 0.474  & 22.2  \\
PD & ADNI-CN &                        & 0.373  &                       & 0.255  & 31.4  \\
AD & PD      & \multirow{2}{*}{1.164} & 0.939  & \multirow{2}{*}{1.091}& 0.770  & 18.0  \\
PD & AD      &                        & 0.620  &                       & 0.434  & 29.9  \\
CN & AD      & \multirow{2}{*}{0.996} & 0.753  & \multirow{2}{*}{0.997}& 0.583  & 22.6  \\
AD & CN      &                        & 0.917  &                       & 0.773  & 15.8  \\
CN & PD      & \multirow{2}{*}{0.249} & 0.593  & \multirow{2}{*}{0.269}& 0.466  & 21.4  \\
PD & CN      &                        & 0.349  &                       & 0.264  & 24.2  \\
\bottomrule
 \end{tabular}
 \small{$\dagger$:Distance between CN\_SI and AD\_SI were normalized to 1.}
\end{table}

%5.2.2
\subsubsection{Visualization of the distribution of the embedding}
\begin{figure*}[t]
\centering
\includegraphics[width=0.8\linewidth]{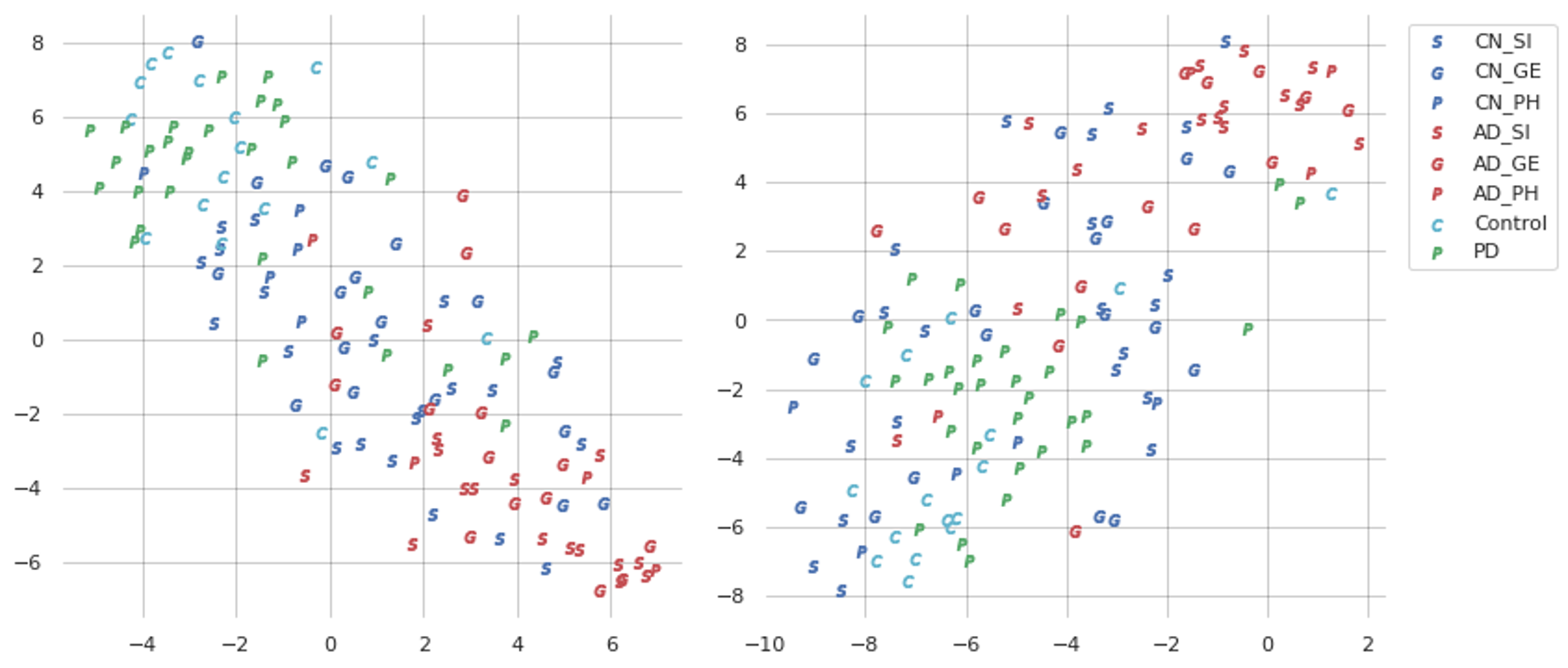}
\caption{Distribution of embedding visualized with t-SNE \cite{Laurens2008}:}
\small{(left)  baseline (3D-CAE+metric learning),  (right) baseline with PSS.}
\label{fig:fig7}
\end{figure*}

Figures \ref{fig:fig7}a and \ref{fig:fig7}b show scatter plots of the embedding of test data with and without PSS, respectively in an arbitrary fold by t-SNE. 
Specifically, this is a scatter plot of the AD, CN, and Control test cases (data excluded from the training in the five-fold cross-validation) along with the untrained PD cases on the model. 
Here, PD has been randomly reduced to 1/5 for better visualization. Without PSS (baseline; 3D CAE + metric learning), AD and CN are properly separated, but the distribution of Control + PD (i.e., the difference in datasets) is separated from that of CN to a discernible degree (left). It can be confirmed that by performing PSS, the distribution of Control + PD becomes closer to that of CN, and the separation between AD and other categories becomes better (right).

%
% Fig 7 (scatter)
%

%5.3
\subsection{Clustering performance of the embedding}
In this section, we compare the separation ability of the obtained low-dimensional embedding of MR images with and without PSS (baseline). 

Tables \ref{table:table5} summarizes the clustering performance evaluated with six commonly used criteria.
These are the silhouette score (silh), homogeneity score (homo), completeness score (comp), V-measure (harmonic mean of homogeneity and completeness; V) , ARI, and AMI implemented on the scikit-learn library. 
In each category, 1 is the best score and 0 is a score based on random clustering. 
It can be confirmed that PSS improved the clustering ability in all evaluation items.

Table \ref{table:table6} is a summary of the clustering performance evaluated with the diagnostic ability. Table \ref{table:table6} (a) is a confusion matrix. Here, the numbers of CN, Control and AD cases are the sum of each fold in the cross-validation. 
In each fold, we tested all PD cases (not included in the training), and the number was divided by five and rounded to the nearest whole number. 
Tables \ref{table:table6}b and \ref{table:table6}c summarize the diagnostic  performance calculated from Table \ref{table:table6} (a) without and with PD cases, respectively. It can be confirmed that PSS enhances the separation of AD and other categories (i.e., CN, Control and PD) in the low-dimensional representation. 

%
% Table 5 (clustering performance)
%
\begin{table}[t]
\begin{center}
\caption{Clustering performance evaluated with common criteria $\dagger$}
\label{table:table5}
\begin{tabular}{crrrrrr}
\toprule
% & silh & homo & comp & V & ARI & AMI \\ \hline
 & \multicolumn{1}{c}{silh} & \multicolumn{1}{c}{homo} & \multicolumn{1}{c}{comp} & \multicolumn{1}{c}{V} & \multicolumn{1}{c}{ARI} & \multicolumn{1}{c}{AMI} \\ \hline
%& silhouette  & homogeneity & Completeness & V-measure & ARI & AMI \\
baseline & 0.236 & 0.220 & 0.301 & 0.250 & 0.251 & 0.241 \\
 +PSS & 0.246 & 0.301 & 0.351 & 0.324 & 0.387 & 0.317 \\
 \bottomrule
 \end{tabular}
 \end{center}
 \small{$\dagger$: Score 1 is the best in each category. 0 is the score for random clustering.}
\end{table}

%
% Table 6 (clustering performance)
%

\begin{table*}[t]
\caption{Evaluation of clustering ability by diagnostic ability.}
\label{table:table6}
% 6a
\begin{subtable}[t]{0.9\textwidth}
\begin{center}
\caption{Confusion matrix}
\begin{tabular}{l|cc|cc}
\toprule
\multirow{2}{*}{}&\multicolumn{2}{c|}{baseline} & \multicolumn{2}{c}{with PSS} \\
                 & CN+Control (+PD)& AD & CN+Control (+PD)& AD    \\ \hline
 CN+Control  &284& 3& 274& 13\\
 (+PD)       &(+104)&(+45)&(+113)&(+36)\\ \hline
 AD          &114 & 66& 75&105 \\
\bottomrule
\end{tabular}
\end{center}
\end{subtable}
%

%6b
\vspace*{0.5em}
\begin{subtable}[t]{0.9\textwidth}
\begin{center}
\caption{Clustering performance (excluded PD cases)}
%\label{table:table6}
\begin{tabular}{l|rrr|rrr|rr}
\toprule
\multirow{2}{*}{}&\multicolumn{3}{c|}{CN+Control} & \multicolumn{3}{c|}{AD}& \multirow{2}{*}{accuracy}&\multirow{2}{*}{ macro-F1} \\
 & precision & recall & \multicolumn{1}{c|}{F1} & precision & recall & \multicolumn{1}{c|}{F1} & & \\ \hline
 baseline & 71.36  & 98.95  & 82.92  & 95.65  & 36.67  & 53.01  & 74.9 & 68.0 \\
+PSS & 78.51  & 95.47  & 86.16  & 88.98  & 58.33  & 70.47  & 81.1 & 78.3 \\
\bottomrule
\end{tabular}
\end{center}
\end{subtable}
%

%6c
\vspace*{0.5em}
\begin{subtable}[t]{0.9\textwidth}
\begin{center}
\caption{Clustering performance (included PD cases)}
%\label{table:table6}
\begin{tabular}{l|rrr|rrr|rrr}
\toprule
\multirow{2}{*}{}&\multicolumn{3}{c|}{CN+Control} & \multicolumn{3}{c|}{AD}& \multirow{2}{*}{accuracy}&\multirow{2}{*}{ macro-F1} & Specificity\\
 & precision & recall & F1 & precision & recall & F1 & & & of PD\\ \hline
baseline & 77.29  & 88.99  & 82.73  & 57.89  & 36.67  & 44.90  & 73.7 & 63.8 & 69.7 \\
+PSS & 83.77  & 88.76  & 86.19  & 68.18  & 58.33  & 62.87  & 79.9 & 74.5 & 75.8 \\
\bottomrule
\end{tabular}
\end{center}
\end{subtable}
\end{table*}

PSS improved the diagnostic performance by about 6.2\% (from 73.7 to 79.9\%) for micro-accuracy and about 10.7\% (from 63.8 to 74.5\%) for  macro-F1. The specificity for PD was also improved by 6.1\% (from 69.7\% to 75.8\%).

%% section 6
\section{Discussion}
\label{sec:sec6}
%6.1
\subsection{Changes on MR images by PSS}
Our PSS network transforms healthy cases taken with GE scanners to those taken with Siemens scanners. 
As can be seen from Figure \ref{fig:fig6} and Table \ref{table:table2}, the amount of change in the images because of PSS was almost the same for both AD and CN images in the ADNI dataset, including the Philips case.

The amount of conversion of the image for the PPMI dataset was larger than that for the ADNI dataset. This is thought to be due to the process of absorbing the differences in the datasets that exist in the image but are invisible to the eye.
However, in all cases, the converted images have a natural appearance without destroying the brain structure. 
This can be objectively confirmed in SSIM, which evaluates the structural similarity on the image, maintains a high value. As discussed in detail below, PSS can reduce disease-specific variation in the resulting low-dimensional embedding, absorb differences among datasets and scanner vendors, and improve the separability of diseases.
Given these factors, we can conclude that this PSS transformation was done properly.

%6.2
\subsection{Contributions of DI-PSS for CBIR}
This section discusses the effects of our DI-PSS framework from the perspective of CBIR implementation.
%6.2.1
\subsubsection{Distribution of embedding}
Based on the results in Tables \ref{table:table3} and \ref{table:table4}, we first discuss the effectiveness of the proposed DI-PSS.
From Table \ref{table:table3}, PSS reduces the inter-cluster variability for all data categories.
In particular, the SD of ADNI-CN and ADNI-AD, which are taken by scanners from three different companies in the same dataset, are reduced by 6.9\% and 6.1\%, respectively.
This indicates that the PSS reduces the difference caused by different scanners.
In addition, the SD of ALL\_CN, which is a combination of ADNI-CN and Control  from a different PPMI dataset, is also reduced by 7.2\%, which clearly shows that the proposed PSS can absorb differences in datasets. 
This benefit can also be seen in Figure \ref{fig:fig7}.

The reduction of PD variability by PSS is more pronounced ($-$14.7\%) than the others, and it is ultimately the category with the lowest variability. This is  mentioned later in this section.
From Table \ref{table:table4}, PSS also succeeds in reducing the variability from each piece of data to all the different cluster centers (inter-cluster variability). What is noteworthy here is the degree of decrease in the standard deviation, which reached an average of 22.6\%.
This ability to reduce not only the variability of data in the same category, but also the directional variability up to different data categories is an important feature in CBIR.

%GE_CN->SI_CNだけでも実現しているのはうれしい
In this experiment, we only built an image transformer (i.e. PSSnetwork) that converts CN\_GE to CN\_SI cases, but we could confirm that the harmonization is desirable for categories that are not included in the training in this way.
This strongly suggests that the strategy we have adopted -- that is, not having to build image harmonizers for all scanner types -- may have sufficient harmonization effects for many types of scanners.

%PDはCNと同じとみなしたことの妥当性とこれについての課題
%それでもPD-CNの距離が増加し、PD自体のばらつきは小さくなっている。
%→潜在的にPDがCNとは違うクラスとして扱えるかも。
Incidentally, the distances between PD and CN (ADNI-CN vs. PD and ALL-CN vs. PD) are closer than the distances between other categories. This supports the validity of the assumption we made in our experiment that PD and CN are outwardly indistinguishable, and therefore, they can be treated as the same class. 
In contrast, if we look closely, we can see that the distances of the gravity centers between PD and CN (0.249$\rightarrow$0.269) and PD and Control (0.256$\rightarrow$0.297) are slightly increased by PSS, and Table \ref{table:table3} shows that the variation of PD is greatly reduced by PSS.
From this, we can say that the PSS is moving the PDs into smaller groups away from CN and Control.
This can be taken as an indication that the model trained by DI-PSS tends to consider PD as a different class that is potentially separated from the CN category.
Since the size of the dataset for this experiment was limited, we would like to run tests with a larger dataset in the future.

%6.2.2
\subsubsection{Separability of the embedding for CBIR}
%DI-PSSはPSS調和のおかげで、距離学習だけではだめだったばらつき低減、性能向上
Thanks to the harmonization of scanners by PSS, the proposed DI-PSS not only reduces the variability of low-dimensional representations of each disease category, which could not be reduced by deep metric learning learning alone as adopted in DCMML \cite{Onga2019}, but also reduces the differences among datasets, resulting in a significant performance improvement in the clustering ability of low-point representations.
% PDの性能があがっていることはCBIRにとって重要
The PD data are different from the ADNI data used for training, and thus, it is an unknown dataset from our model. The improvement of clustering performance by the proposed DI-PSS for PD as well is an important and noteworthy result for the realization of CBIR.

%6.3
\subsection{Validity of the model architecture}
The recently proposed data harmonization methods for brain MR images by Moyer et al \cite{Moyer2020} and Dinsdale et al. \cite{Dinsdale2021} have been reported to be not only logically justified but also very effective. 
However, as mentioned above, these methods are difficult to apply to CBIR applications because images from all scanners are theoretically needed to train the model. Our DI-PSS is a new proposal to address these problems.

Although DI-PSS only learned the transformation from CN\_GE to CN\_Siemens, the improvement of the properties of the obtained embeddings was confirmed even for combinations that included other companies' scanners, such as the Philips scanner, and different disease categories (AD) that were not included in the training.
The results are evidence of proper data harmonization. We think this is due to the combination of MRICloud, an advanced skull stripping algorithm that performs geometric and volumetric positioning, and CycleGAN's generic style transformation capabilities and distance metric learning, which make up the PSS network.
Experiments with large-scale data from more diverse disease classes are needed, but in this experiment, we could confirm the possibility of obtaining effective scanner standardization by building one model that translates into a standard scanner.

\section*{Limitations of this study}
The number of data and diversity of their conditions used in these experiments are limited. There is also a limit to the number of diseases we considered. In the future, verification using more data is essential.

\section{Conclusion}
In this paper, we proposed a novel and effective MR image embedding method, DI-PSS, which is intended for application to CBIR. DI-PSS achieves data harmonization by transforming MR images to look like those captured with a predefined standard scanner, reducing the bias caused by variations in scanners and scan protocols, and obtaining a low-dimensional representation preserving disease-related anatomical features. The DI-PSS did not require training data that contained MRIs from all scanners and protocols; One set of image converters (i.e., CN\_GE to CN\_Siemens) was sufficient to train the model. In the future, we will continue the validation with more extensive and diverse data.

\section*{Acknowledgment}
 This research was supported in part by the Ministry of Education, Science, Sports and Culture of Japan (JSPS KAKENHI), Grant-in-Aid for Scientific Research (C), 21K12656, 2021–2023.
\begin{footnotesize}
The MRI data collection and sharing for this project was funded by the Alzheimer’s Disease Neuroimaging Initiative (ADNI) (National Institutes of Health Grant U01 AG024904) and DOD ADNI (Department of Defense award number W81XWH-12–2-0012). 
ADNI is funded by the National Institute on Aging, the National Institute of Biomedical Imaging and Bioengineering, and through generous contributions from the following: AbbVie, Alzheimer’s Association; Alzheimer’s Drug Discovery Foundation; Araclon Biotech; BioClinica, Inc.; Biogen; Bristol-Myers Squibb Company; CereSpir, Inc.; Cogstate; Eisai Inc.; Elan Pharmaceuticals, Inc.; Eli Lilly and Company; EuroImmun; F. Hoffmann-La Roche Ltd and its affiliated company Genentech, Inc.; Fujirebio; GE Healthcare; IXICO Ltd.; Janssen Alzheimer Immunotherapy Research \& Development, LLC.; Johnson \& Johnson Pharmaceutical Research \& Development LLC.; Lumosity; Lundbeck; Merck \& Co., Inc.; Meso Scale Diagnostics, LLC.; NeuroRx Research; Neurotrack Technologies; Novartis Pharmaceuticals Corporation; Pfizer Inc.; Piramal Imaging; Servier; Takeda Pharmaceutical Company; and Transition Therapeutics.
The Canadian Institutes of Health Research is providing funds to support ADNI clinical sites in Canada. Private sector contributions are facilitated by the Foundation for the National Institutes of Health (www.fnih.org). The grantee organization is the Northern California Institute for Research and Education, and the study is coordinated by the Alzheimer’s Therapeutic Research Institute at the University of Southern California. 
ADNI data are disseminated by the Laboratory for Neuro Imaging at the University of Southern California.
An additional MRI data used in the preparation of this article were obtained from the Parkinson’s Progression Markers Initiative (PPMI) database (\url{www.ppmi-info.org/data}).
For up-to-date information on the study, visit www.ppmi-info.org. PPMI – a public-private partnership – is funded by the Michael J. Fox Foundation for Parkinson’s Research and funding partners, including AbbVie, Allergan, Avid Radiopharmaceuticals, Biogen, Biolegend, Bristol-Myers Squibb, Celgene, Denali, GE Healthcare, Genentech, GlaxoSmithKline, Lilly, Lundbeck, Merck, Meso Scale Discovery, Pfizer, Piramal, Prevail Therapeutics, Roche, Sanofi Genzyme, Servier, Takeda, Teva, UCB, Verily, Voyager Therapeutics, and Golub Capital.
\end{footnotesize}

%\appendices
%Appendixes, if needed, appear before the acknowledgment.

%\bibliographystyle{ieeeaccess}
%\bibliographystyle{unsrt}
\bibliographystyle{IEEEtran}
\bibliography{references}

\begin{thebibliography}{10}
\providecommand{\url}[1]{#1}
\csname url@samestyle\endcsname
\providecommand{\newblock}{\relax}
\providecommand{\bibinfo}[2]{#2}
\providecommand{\BIBentrySTDinterwordspacing}{\spaceskip=0pt\relax}
\providecommand{\BIBentryALTinterwordstretchfactor}{4}
\providecommand{\BIBentryALTinterwordspacing}{\spaceskip=\fontdimen2\font plus
\BIBentryALTinterwordstretchfactor\fontdimen3\font minus
  \fontdimen4\font\relax}
\providecommand{\BIBforeignlanguage}[2]{{%
\expandafter\ifx\csname l@#1\endcsname\relax
\typeout{** WARNING: IEEEtran.bst: No hyphenation pattern has been}%
\typeout{** loaded for the language `#1'. Using the pattern for}%
\typeout{** the default language instead.}%
\else
\language=\csname l@#1\endcsname
\fi
#2}}
\providecommand{\BIBdecl}{\relax}
\BIBdecl

\bibitem{Woelfe2011}
M.~Woelfle, P.~Olliaro, and M.~H. Todd, ``Open science is a research
  accelerator,'' \emph{Nature chemistry}, vol.~3, no.~10, pp. 745--748, 2011.

\bibitem{Kumar2013}
A.~Kumar, J.~Kim, W.~Cai, M.~Fulham, and D.~Feng, ``Content-based medical image
  retrieval: a survey of applications to multidimensional and multimodality
  data,'' \emph{Journal of Digital Imaging}, vol.~26, no.~6, pp. 1025--1039,
  2013.

\bibitem{Tu2010}
Z.~Tu and X.~Bai, ``Auto-context and its application to high-level vision tasks
  and 3{D} brain image segmentation,'' \emph{IEEE Transactions on Pattern
  Analysis and Machine Intelligence}, vol.~32, no.~10, pp. 1744--1757, 2010.

\bibitem{Huang2012}
M.~Huang, W.~Yang, M.~Yu, Z.~Lu, Q.~Feng, and W.~Chen, ``Retrieval of brain
  tumors with region-specific bag-of-visual-words representations in
  contrast-enhanced {MRI} images,'' \emph{Computational and Mathematical
  Methods in Medicine}, vol. 2012, p. 280538, 2012.

\bibitem{Murala2014}
S.~Murala and Q.~M.~J. Wu, ``Local mesh patterns versus local binary patterns:
  Biomedical image indexing and retrieval,'' \emph{IEEE Journal of Biomedical
  and Health Informatics}, vol.~18, no.~3, pp. 929--938, 2014.

\bibitem{Faria2015}
A.~V. Faria, K.~Oishi, S.~Yoshida, A.~Hillis, M.~I. Miller, and S.~Mori,
  ``Content-based image retrieval for brain {MRI}: An image-searching engine
  and population-based analysis to utilize past clinical data for future
  diagnosis,'' \emph{NeuroImage: Clinical}, vol.~7, pp. 367--376, 2015.

\bibitem{Kruthika2019}
K.~Kruthika, Rajeswari, and H.~Maheshappa, ``{CBIR} system using capsule
  networks and 3{D} {CNN} for alzheimer's disease diagnosis,''
  \emph{Informatics in Medicine Unlocked}, vol.~14, pp. 59--68, 2019.

\bibitem{Swati2019}
Z.~N.~K. Swati, Q.~Zhao, M.~Kabir, F.~Ali, Z.~Ali, S.~Ahmed, and J.~Lu,
  ``Content-based brain tumor retrieval for {MR} images using transfer
  learning,'' \emph{IEEE Access}, vol.~7, pp. 17\,809--17\,822, 2019.

\bibitem{Onga2019}
Y.~Onga, S.~Fujiyama, H.~Arai, Y.~Chayama, H.~Iyatomi, and K.~Oishi,
  ``Efficient feature embedding of 3{D} brain {MRI} images for content-based
  image retrieval with deep metric learning,'' pp. 3764--3769, 2019.

\bibitem{Alipanahi2008}
B.~Alipanahi, M.~Biggs, and A.~Ghodsi, ``Distance metric learning vs. fisher
  discriminant analysis,'' \emph{Proceedings of the 23rd National Conference on
  Artificial Intelligence}, vol.~2, pp. 598--603, 2008.

\bibitem{Hoffer2016}
E.~Hoffer and N.~Ailon, ``Semi-supervised deep learning by metric embedding,''
  \emph{arXiv preprint}, p. 1611.01449, 2016.

\bibitem{Clark2006}
K.~A. Clark, R.~P. Woods, D.~A. Rottenberg, A.~W. Toga, and J.~C. Mazziotta,
  ``Impact of acquisition protocols and processing streams on tissue
  segmentation of t1 weighted mr images,'' \emph{NeuroImage}, vol.~29, no.~1,
  pp. 185--202, 2006.

\bibitem{Han2006}
X.~Han, J.~Jovicich, D.~Salat, A.~van~der Kouwe, B.~Quinn, S.~Czanner, E.~Busa,
  J.~Pacheco, M.~Albert, R.~Killiany, P.~Maguire, D.~Rosas, N.~Makris, A.~Dale,
  B.~Dickerson, and B.~Fischl, ``Reliability of {MRI}-derived measurements of
  human cerebral cortical thickness: the effects of field strength, scanner
  upgrade and manufacturer,'' \emph{Neuroimage}, vol.~32, no.~1, pp. 180--194,
  2006.

\bibitem{Yu2018}
M.~Yu, K.~A. Linn, P.~A. Cook, M.~L. Phillips, M.~McInnis, M.~Fava, M.~H.
  Trivedi, M.~H. Trivedi, R.~T. Shinohara, and Y.~I. Sheline, ``Statistical
  harmonization corrects site effects in fuctional connectivity measures from
  multi-site f{MRI} data,'' \emph{Human brain mapping}, vol.~39, no.~11, pp.
  4213--4227, 2018.

\bibitem{Oishi2019}
K.~Oishi, J.~Chotiyanonta, D.~Wu, M.~I. Miller, and S.~Mori, ``Developmental
  trajectories of the human embryologic brain regions,'' \emph{Neuroscience
  Letters}, vol. 708, p. 134342, 2019.

\bibitem{Wachinger2021}
C.~Wachinger, A.~Rieckmann, and S.~Pölsterl, ``Detect and correct bias in
  multi-site neuroimaging datasets,'' \emph{Medical Image Analysis}, vol.~67,
  p. 101879, 2021.

\bibitem{Gao2019}
Y.~Gao, J.~Pan, Y.~Guo, J.~Yu, J.~Zhang, D.~Geng, and Y.~Wang, ``Optimised mri
  intensity standardisation based on multi-dimensional sub-regional point cloud
  registration,'' \emph{Computer Methods in Biomechanics and Biomedical
  Engineering: Imaging \& Visualization}, vol.~7, no. 5-6, pp. 594--603, 2019.

\bibitem{Um2019}
H.~Um, F.~Tixier, D.~Deasy, Bermudez, J.~O, R.~J. Young, and H.~Veeraraghavan,
  ``Impact of image preprocessing on the scanner dependence of multi-parametric
  {MRI} radiomic features and covariate shift in multi-institutional
  glioblastoma datasets,'' \emph{Physics in Medicine and Biology}, vol.~64,
  no.~16, p. 165011, 2019.

\bibitem{Johnson2007}
W.~E. Johnson, C.~Li, and A.~Rabinovic, ``{Adjusting batch effects in
  microarray expression data using empirical Bayes methods},''
  \emph{Biostatistics}, vol.~8, no.~1, pp. 118--127, 2006.

\bibitem{Fortin2017-COMBAT}
\BIBentryALTinterwordspacing
J.-P. Fortin, D.~Parker, B.~Tunc, T.~Watanabe, M.~A. Elliott, K.~Ruparel, D.~R.
  Roalf, T.~D. Satterthwaite, R.~C. Gur, R.~E. Gur, R.~T. Schultz, R.~Verma,
  and R.~T. Shinohara, ``Harmonization of multi-site diffusion tensor imaging
  data,'' \emph{bioRxiv}, 2017. [Online]. Available:
  \url{http://biorxiv.org/content/early/2017/03/15/116541}
\BIBentrySTDinterwordspacing

\bibitem{Fortin2017a}
J.-P. Fortin, D.~Parker, B.~Tunç, T.~Watanabe, M.~A. Elliott, K.~Ruparel,
  D.~R. Roalf, T.~D. Satterthwaite, R.~C. Gur, R.~E. Gur, R.~T. Schultz,
  R.~Verma, and R.~T. Shinohara, ``Harmonization of multi-site diffusion tensor
  imaging data,'' \emph{NeuroImage}, vol. 161, pp. 149--170, 2017.

\bibitem{Fortin2017b}
J.-P. Fortin, N.~Cullen, Y.~I. Sheline, W.~D. Taylor, I.~Aselcioglu, P.~A.
  Cook, P.~Adams, C.~Cooper, M.~Fava, P.~J. McGrath, M.~McInnis, M.~L.
  Phillips, M.~H. Trivedi, and M.~M. Weissman, ``Harmonization of cortical
  thickness measurements across scanners and sites,'' \emph{NeuroImage}, vol.
  167, pp. 104--120, 2017.

\bibitem{Zhao2019}
F.~Zhao, Z.~Wu, L.~Wang, W.~Lin, S.~Xia, D.~Shen, and G.~Li, ``Harmonization of
  infant cortical thickness using surface-to-surface cycle-consistent
  adversarial networks,'' \emph{Med Image Comput Comput Assist Interv}, vol.
  11767, pp. 475--483, 2019.

\bibitem{Dewey2019}
C.~Zhao, J.~C. Reinhold, A.~Carass, K.~C. Fitzgerald, E.~S. Sotirchos,
  S.~Saidha, J.~Oh, D.~L. Pham, P.~A. Calabresi, P.~C.~M. van Zijl, and J.~L.
  Prince, ``Deepharmony: A deep learning approach to contrast harmonization
  across scanner changes,'' \emph{Magnetic Resonance Imaging}, vol.~64, pp.
  160--170, 2019.

\bibitem{Moyer2020}
D.~Moyer, G.~Ver~Steeg, C.~M.~W. Tax, and T.~P. M., ``Scanner invariant
  representations for diffusion {MRI} harmonization,'' \emph{Magnetic resonance
  in medicine}, vol.~84, no.~4, pp. 2174--2189, 2020.

\bibitem{Dinsdale2021}
N.~K. Dinsdale, M.~Jenkinson, and A.~I.~L. Namburete, ``Deep learning-based
  unlearning of dataset bias for {MRI} harmonisation and confound removal,''
  \emph{NeuroImage}, vol. 228, p. 117689, 2021.

\bibitem{Zhu2017}
J.-Y. Zhu, T.~Park, P.~Isola, and A.~A. Efros, ``Unpaired image-to-image
  translation using cycle-consistent adversarial networks,'' pp. 2242--2251,
  2017.

\bibitem{Ganin2016}
Y.~Ganin, E.~Ustinova, H.~Ajakan, P.~Germain, H.~Larochelle, F.~Laviolette,
  M.~Marchand, and V.~Lempitsky, ``Domain-adversarial training of neural
  networks,'' \emph{Journal of Machine Learning Research}, vol.~17, pp. 1--35,
  2016.

\bibitem{Laurens2008}
M.~van~der Laurens and G.~Hinton, ``Visualizing data using t-{SNE},''
  \emph{Journal of machine learning research}, vol.~9, no.~86, pp. 2579--2605,
  2008.

\bibitem{Mori2016}
S.~Mori, D.~Wu, C.~Ceritoglu, Y.~Li, A.~Kolasny, M.~A. Vaillant, A.~V. Faria,
  K.~Oishi, and M.~I. Miller, ``{MRIC}loud: Delivering high-throughput mri
  neuroinformatics as cloud-based software as a service,'' \emph{Computing in
  Science Engineering}, vol.~18, no.~5, pp. 21--35, 2016.

\bibitem{Arai2018}
H.~Arai, Y.~Chayama, H.~Iyatomi, and K.~Oishi, ``Significant dimension
  reduction of 3{D} brain {MRI} using 3{D} convolutional autoencoders,'' pp.
  5162--5165, 2018.

\bibitem{Postuma2015}
R.~B. Postuma, D.~Berg, M.~Stern, W.~Poewe, C.~W. Olanow, W.~Oertel, J.~Obeso,
  K.~Marek, I.~Litvan, A.~E. Lang, G.~Halliday, C.~G. Goetz, T.~Gasser,
  B.~Dubois, P.~Chan, B.~R. Bloem, C.~H. Adler, and G.~Deuschl, ``{MDS}
  clinical diagnostic criteria for {P}arkinson's disease,'' \emph{Movement
  Disorders}, vol.~30, no.~12, pp. 1591--601, 2015.

\bibitem{Meijer2017}
F.~J. Meijera, B.~Goraja, B.~R. Bloemc, and R.~A. Esselinkc, ``Clinical
  application of brain mri in the diagnostic work-up of parkinsonism,''
  \emph{Journal of Parkinson’s Disease}, vol.~7, pp. 211--217, 2017.

\bibitem{Oishi2009}
K.~Oishi, A.~Faria, H.~Jiang, X.~Li, K.~Akhter, J.~Zhang, J.~T. Hsu, M.~I.
  Miller, P.~C.~M. van Zijl, M.~Albert, C.~G. Lyketsos, R.~Woods, A.~W. Toga,
  G.~B. Pike, P.~Rosa~Neto, A.~Evans, J.~Mazziotta, and S.~Mori, ``Atlas-based
  whole brain white matter analysis using large deformation diffeomorphic
  metric mapping: application to normal elderly and alzheimer's disease
  participants,'' \emph{Neuroimage}, vol.~46, no.~2, pp. 486--499, 2009.

\bibitem{Ng2001}
A.~Y. Ng, M.~I. Jordan, and Y.~Weiss, ``On spectral clustering: Analysis and an
  algorithm,'' \emph{Proceedings of the 14th International Conference on Neural
  Information Processing Systems: Natural and Synthetic}, pp. 849--856, 2001.

\end{thebibliography}

% author's bios are excluded 
\EOD
\end{document}